# New pyramidal hybrid textural and deep features based automatic skin cancer classification model: Ensemble DarkNet and textural feature extractor

Mehmet Baygin[a], Turker TUNCER[b], Sengul DOGAN[b]

[a] Department of Computer Engineering, College of Engineering, Ardahan University, Ardahan, Turkey

mehmetbaygin@ardahan.edu.tr

[b] Department of Digital Forensics Engineering, Technology Faculty, Firat University, Elazig, Turkey

{sdogan, turkertuncer}@firat.edu.tr

**Abstract**

*Background:* Skin cancer is one of the widely seen cancer worldwide and automatic classification of skin cancer can be benefited dermatology clinics for an accurate diagnosis. Hence, a machine learning-based automatic skin cancer detection model must be developed.

*Material and Method:* This research interests to overcome automatic skin cancer detection problem. A colored skin cancer image dataset is used. This dataset contains 3297 images with two classes. An automatic multilevel textural and deep features-based model is presented. Multilevel fuse feature generation using discrete wavelet transform (DWT), local phase quantization (LPQ), local binary pattern (LBP), pre-trained DarkNet19, and DarkNet53 are utilized to generate features of the skin cancer images, top 1000 features are selected threshold value-based neighborhood component analysis (NCA). The chosen top 1000 features are classified using the 10-fold cross-validation technique.

*Results:* To obtain results, ten-fold cross-validation is used and 91.54% classification accuracy results are obtained by using the recommended pyramidal hybrid feature generator and NCA selector-based model. Further, various training and testing separation ratios (90:10, 80:20, 70:30, 60:40, 50:50) are used and the maximum classification rate is calculated as 95.74% using the 90:10 separation ratio.

*Conclusions:* The findings and accuracies calculated are denoted that this model can be used in dermatology and pathology clinics to simplify the skin cancer detection process and help physicians.

**Keywords:** Pyramidal hybrid feature generation, ensemble pre-trained DarkNet feature extractor; textural features; NCA; skin cancer detection.

## 1. Introduction

Dermatological diseases are one of the most important health problems encountered today. In these diseases, one of the most dangerous is skin cancer [1,2]. Cancer as a term means cells that divide and grow uncontrolled [3]. At this point, one of the most common types of cancer today is skin cancer [4,5]. Recent studies reveal that both non-melanoma and melanoma skin cancers have increased significantly in the last decade [5,6]. Again, according to these studies, approximately 2-3 million non-melanomas and 132,000 melanoma skin cancers are diagnosed every year [7,8]. One of the most important reasons for this situation is that the function of the atmosphere to absorb harmful sunlight is decreasing day by day [8]. Today, deterioration in the ozone layer causes an increase in the number of skin cancer cases [9].

The diagnosis of skin cancer is usually made by expert dermatologists by analyzing dermoscopic images, as well as biopsy and histopathological examination [10]. However, these processes are both time-consuming and highly susceptible to human error. Studies in the literature show that skin cancer cases can be largely treated with early diagnosis [9,11]. Therefore, fast and accurate classification of dermoscopic images is a very important issue [12,13]. Nowadays, many diseases can be automatically pre-diagnosed with artificial intelligence technology [2,3,14,15]. In this paper, a new classification model has been proposed to make a pre-diagnosis of skin cancer cases. Details of this proposed method are described in the following sections.

### 1.1. Literature Review

Artificial intelligence technologies are one of the hot topics that are frequently studied in the literature on medical diagnosis. In this study, a new classification method that can make automatic and rapid pre-diagnosis of skin cancer cases is proposed. Details of some studies conducted on this subject in the literature are given in Table 1.

Table 1. Studies on automatic skin cancer detection

| Author(s) and Year | Method | Dataset | Validation Method | Classifier | Results |
|---|---|---|---|---|---|
| J. Amin et. al. [16], 2020 | 2D Wavelet transform Otsu algorithm | PH2 [17] ISBI 2016 [18] | 5 fold-cv 50:50 hold-out cv | Ensemble KNN | 5-fold cv Acc.=98.71% |

| | Deep feature generation (AlexNet+VGG16) | ISBI 2017 [18] | | Discriminant Analysis SVM Tree | 50:50 Split Acc.=99.9% |
|---|---|---|---|---|---|
| N. Zhang et. al. [19], 2020 | CNN Whale optimization algorithm | DermQuest DermIS [20] | 80:20 | CNN | Acc.~=95% |
| A. Hekler et. al. [10], 2019 | Doctor's examination CNN | HAM10000 [21] | --- | CNN | Acc.=81.59% |
| M. Togacar et. al. [22], 2021 | Autoencoder MobileNetV2 Spiking Neural Networks | ISIC [23] | 10 fold-cv | SVM | Acc.=95.27% |
| V. Srividhya et. al. [24], 2020 | Highly Perceptive Features CNN | DermQuest DermIS | --- | CNN | Acc.=95% |
| M. K. Monika et. al. [25], 2020 | Statistical and texture feature extraction | ISIC 2019 [26] | 70:30 | SVM | Acc.=96.25% |
| M. A. Kadampur et. al. [27], 2020 | Deep learning models ResNet, SqueezeNet, DenseNet, InceptionV3 | HAM10000 | 80:10:10 | Deep learning classifiers | InceptionV3 Pre.=98.19% F1=95.74% AUC=99.23% |
| M. Elgamal [28], 2013 | DWT PCA | Online Collection (Total 40 | n fold-cv | ANN KNN | ANN Acc.=95% Sen.=95% Spe.=95% |

| | | | | | KNN<br>Acc.=97.5%<br>Sen.=100%<br>Spe.=95% |
|---|---|---|---|---|---|
| M. Koklu et. al. [29], 2017 | ABCD rule criteria | PH2 | 10 fold-cv | ANN<br>SVM<br>KNN<br>DT | ANN<br>Acc.=92.50%<br>SVM<br>Acc.=89.50%<br>KNN<br>Acc.=82.00%<br>DT<br>Acc.=90.00% |
| R. Ashraf et. al. [30], 2020 | ROI extract<br>CNN | DermQuest<br>DermIS | 77:23 | CNN | DermQuest<br>Acc.=97.4%<br>DermIS<br>Acc.=97.9% |
| M. Vidya et. al. [31], 2020 | ABCD rule<br>GLCM<br>HOG | ISIC | 80:20 | SVM<br>KNN<br>Naïve Bayes | SVM<br>Acc.=97.8<br>KNN<br>Acc.=95%<br>Naïve Bayes<br>Acc.=91.2% |
| S. Jinnai et. al. [32], 2020 | CNN | Own | --- | CNN | Two class<br>Acc.=91.5%<br>Six class<br>Acc.=86.2% |

## 1.2. Contributions

Contributions of the recommended pyramidal hybrid feature generation and NCA selector based

- New pyramidal feature generator is presented. The main aim of this generator is to generate both textural and deep features. Textural features are hand-crafted and they can extract low-level features. To provide low-level and high-level features using textural generator, pyramidal structure is presented. Furthermore, pre-trained both DarkNet19 and DarkNet53 networks are used for feature generation

- Skin cancer classification using images is one of the hard problem for computer vision. Therefore, variable models have been introduced for automatic detection skin cancer. A new skin cancer image classification model is recommended using our pyramidal hybrid feature generator. This model is tested on a dataset and it reached high classification accuracy. Moreover, six validation techniques are employed and the calculated all results are obtained greater than 90% classification rates.

## 2. Materials and Methods

In this section, details of a novel pyramidal hybrid feature generator and NCA based skin cancer classification and utilized skin cancer image dataset are explained. Summarization of the presented model is shown in Fig. 1 schematically.

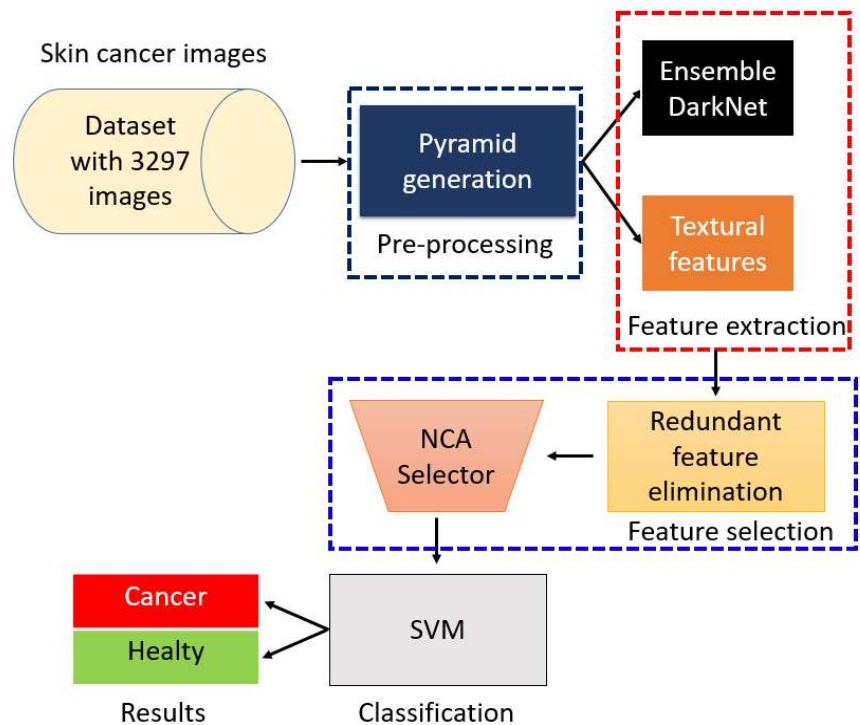

Fig. 1. Snapshot of the proposed pyramidal hybrid feature generator and NCA selector based image classification model.

### 2.1. Material

In this study, a publicly shared dataset is used [23]. This dataset contains images of benign and malignant skin moles and has a balanced structure. The dataset consists of 3297 images in total. The dataset includes 1800 benign and 1497 malignant tumor images. All rights of the data contained in this dataset shared publicly are subject to International Skin Imaging Collaboration (ISIC) archive rights [33]. In the publicly shared dataset, the images are divided into 2 groups as train (75%) and test (25%). In this study, the train and test set were combined. In the classification process, six different validation techniques were tested using SVM. Some basic features of the dataset are given in Table 2, and some sample images of benign and malignant skin moles are presented in Fig. 2.

Table 2. Features of the datasets

| Feature | Value |
|---|---|
| Image Format | JPEG |
| Image Color | RGB |
| Resolution | 224x224 |
| Image Bit Depth | 24 |
| Total Images | 3297 (1800 benign images and 1497 malign images) |
| Classes | 0 = Benign, 1 = Malign |

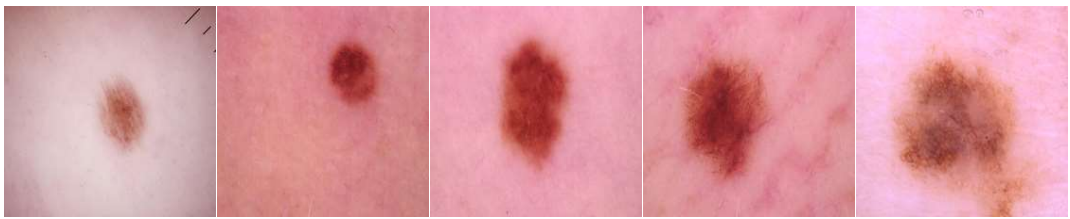

(a) Benign images

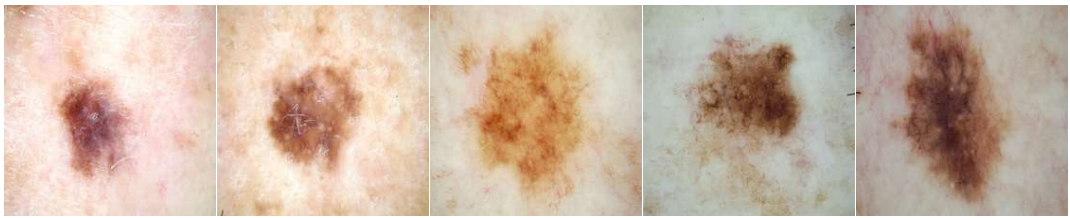

(b) Malign images

Fig. 2. Sample images from dataset

## 2.2. Preliminaries

This research uses two types DarkNet to create ensemble DarkNet deep feature generator. Further, LBP and LPQ textural generators are used for feature creation. The explanations of these models are given in this section briefly.

### 2.2.1. DarkNet

One of the methods used in this study is deep feature extraction. DarkNet deep network architecture is used in the deep feature extraction process. DarkNet architecture has been trained with ImageNet dataset with approximately 1 million images. There are two types: DarkNet19 and DarkNet53. DarkNet19 and DarkNet53 deep network architectures consist of 64 and 184 layers, respectively. In this study, ensemble deep feature generator was constructed using both types of DarkNet. Feature generation has been made from pre-trained DarkNet19 and DarkNet53 deep network architectures using the transfer learning method. In these architectures, "avg1" and "conv53" layers are used, respectively. A block diagram of Ensemble DarkNet deep feature extraction is given in Fig. 3.

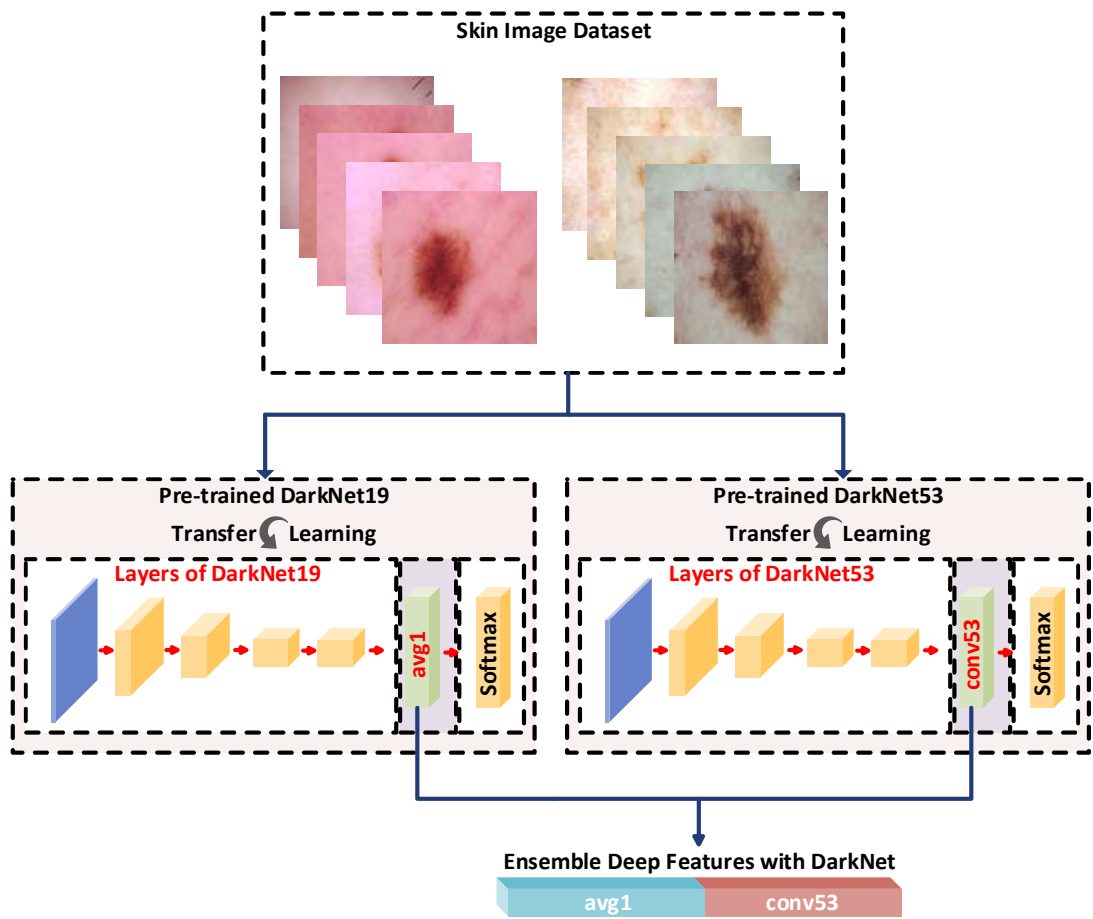

Fig. 3. Ensemble DarkNet feature generation

## 2.2.2. Local binary pattern

In this paper, besides DarkNet-based deep feature extraction, textural feature extraction process is also applied. One of these procedures is Local Binary Pattern (LBP), which is a well-known method in the literature. LBP was first introduced by Ojala in 1994. It is a very effective and simple method. Its computational complexity is very low. It basically starts with the process of splitting the image into 3x3 overlapping blocks [34]. Then, the relationship between the central pixel of each block and its neighboring pixels is determined. Neighboring pixels with a higher numerical value than the central pixel is marked with 1, and neighboring pixels with a lower value are marked with 0. After this process, the pixels obtained in binary are converted into decimals. Mathematical equivalents of these operations are given in Equations 1 and 2. An LBP applied image is constructed by using these values. Finally, the histogram of the LBP applied image is extracted and these histograms are used as features [35]. The pseudo code of the LBP procedure is given in Algorithm 1.

$$bit(k) = signum(P_N, P_C) = \begin{cases} 0, P_N < P_C \\ 1, P_N \geq P_C \end{cases} \quad (1)$$

$$value(m,t) = \sum_{k=1}^{8} bit(k) * 2^{k-1}, m = \{1,2,\dots,W-2\}, t = \{1,2,\dots,H-2\} \quad (2)$$

Algorithm 1. LBP feature extraction procedure

| **Input:** Skin Images (SI) with size of WxH |
| **Output:** Feature (feat) with size of 59 |

| 1: | Load SI |
| 2: | **for** i=1 to W-2 **do** |
| 3: |   **for** j=1 to H-2 **do** |
| 4: |     $block = SI(i:i+2, j:j+2)$; // 3x3 sized block division |
| 5: |     $Apply\ Eq.\ 1\ and\ 2$ |
| 6: |   **end for j** |
| 7: | **end for i** |
| 8: | Extract histogram of the value // Here, 256 features are obtained. However, in uniform LBP mapping, 59 features are obtained because each uniform pattern is assigned to a label. |

In the normal LBP procedure, the histogram of the LBP applied image is extracted. Since there are 8 neighbor pixels in total, 256 features are obtained. However, in the uniform LBP mapping process, the transition from 0 to 1 and from 1 to 0 is controlled. In this way, a total of 58 uniform patterns are obtained. In addition, non-uniform patterns are assigned to a single label in this approach. Thus, a total of 59 patterns are obtained, including 58 uniform and 1 non-uniform pattern. These patterns are used as feature vectors. The size of the feature matrix to be obtained according to the number of neighboring pixels is given in Equation 3 [36]. In Eq. 3, $SoF$ is the size of the feature matrix, and $N$ is the total number of neighboring pixels. In addition, a block diagram summarizing this process is presented in Fig. 4.

$$SoF = N * (N - 1) + 3 \quad (3)$$

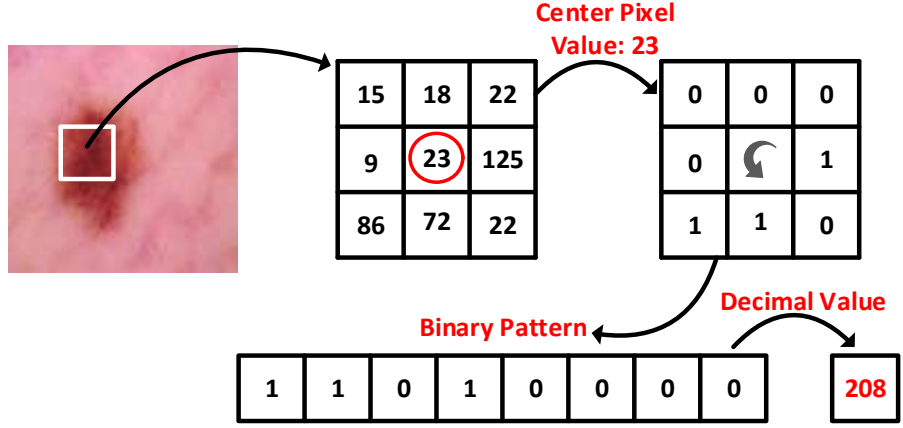

Fig. 4. Flowchart of the LBP procedure

### 2.2.3. Local phase quantization

Another method used in the textural feature extraction process is Local Phase Quntization (LPQ). This process is used to characterize the image texture [37]. It provides feature extraction from the image as blur-insensitive [38,39]. The first step in the LPQ process is to apply a Short Term Fourier Transform (STFT) to the image [40]. This process is given in equation 4.

$$G(u,v) = \sum_{x \in N_x} \sum_{y \in N_y} g(x,y) e^{\frac{-j2\pi(ux+vy)}{M}} \quad (4)$$

Herein, $g(x,y)$ is blurred image, $G(u,v)$ is the Fourier transform of the blurred image, $N_x$ and $N_y$ are neighbor regions, $M$ is the neighbors in the center pixel ($MxM$). LPQ extracts phase information at frequency points $u1 = (a,0), u2 = (0,a), u3 = (a,a)\ and\ u4 = (a,-a)$ using STFT operation. The results obtained for each frequency point are combined and divided into real and imaginary parts. These operations are given in equations 5 and 6.

$$V = [G(u_1), G(u_2), G(u_3), G(u_4)] \quad (5)$$
$$W = [Re\{V\}, Im\{V\}] \quad (6)$$

where, $W$ contains the real ($Re$) and imaginary ($Im$) parts of the phase information. After this process, the elements in $W$ are quantized as follows:

$$q_i = \begin{cases} 1, & W_i \geq 0 \\ 0, & other \end{cases} \quad (7)$$

Herein, the real and imaginary parts are marked as 1 if positive and 0 if negative. This process is the quantization ($q_i$) of the $i\ th$ element. Then, $q_i$ values are encoded using equation 8 and blur-insensitive textural information is obtained.

$$b = \sum_{i=1}^{8} q_i 2^{i-1} \tag{8}$$

Herein, $b$ values are integer values between 0-255. After this process, the histogram of the $b$ values is extracted and used as the feature vector in the classification process.

### 2.3. The proposed pyramidal hybrid feature generation and NCA selector based skin image classification model

The presented model has four primary phases and they are preprocessing (image decomposition for pyramidal model creation), ensemble DarkNet and texture based feature creation, the top informative features selection eliminating redundant features and employing NCA selector and classification utilizing SVM. DWT creates the pyramidal structure. Here, low pass filter sub-bands of the images are used to create pyramid. Therefore, In the proposed feature generation model, ensemble DarkNet extractor is presented and generated 2000 features from each images. In the ensemble DarkNet feature generation model pre-trained DarkNet19 and DarkNet53 are deployed. Here, these networks are trained deploying 1000 classes of the ImageNet dataset. Thus, each network generates 1000 features. Moreover, the widely used textural extractors, are LPQ and LBP are deployed to extract hand-crafted features. Afterward the used deep and textural generators are employed to the created compressed images and raw image and the generated features are merged. The generated and merged features are normalized using min-max normalization and the redundant features are eliminated using a basic process. The top 512 features are selected using the NCA selector and classification results are calculated using SVM. Snapshot of the presented model is shown in Fig. 5.

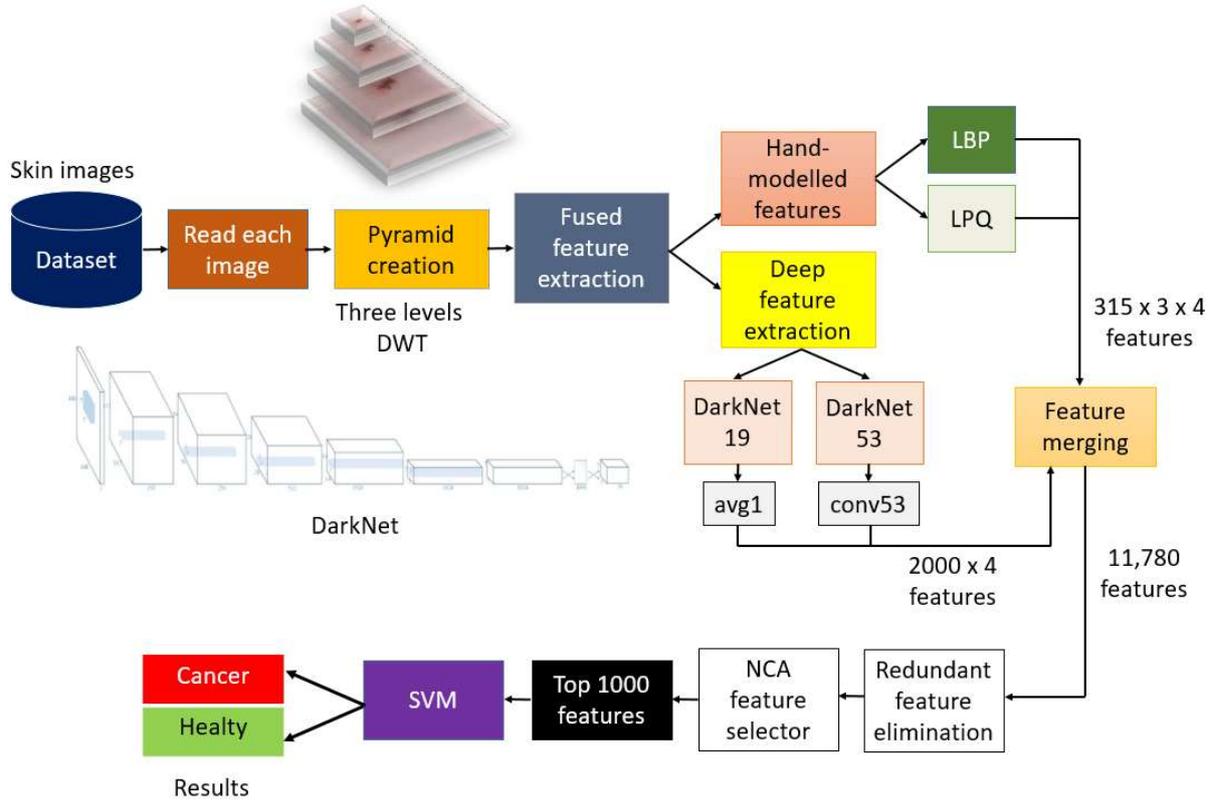

Fig. 5. Schematic explanation of the presented pyramidal ensemble DarkNet and textural features generation based skin cancer recognition model.

Furthermore, the pseudocode of the pyramidal ensemble DarkNet and textural features generation based skin cancer recognition are given below for summarization.

Algorithm 2. The pyramidal ensemble DarkNet and textural features extractor and NCA selector based skin image classification model.

| **Input:** Skin image dataset ($SID$) with 3297 color images. The size of each image is 224 x 224 x 3. |
| --- |
| **Output:** Results. |
| 00: Load the used dataset. |
| 01: **for** t=1 to 3297 **do** |
| 02:     $I = SID(t)$; // Image reading from dataset |
| 03:     **for** h=1 to 4 **do** |
| 04:         $F^h(1:2945) = conc(DarkNet19(I), DarkNet53(I), LBP(I), LPQ(I))$; |
| // In Line 04, $conc$ and $F^h$ are feature merging function and h$^{th}$ generated features respectively. LBP and LPQ generates features from each layer of the used RGB image. |
| 05:         $[LL, LH, HL, HH] = DWT2(I, 'db4')$; // Deploy DWT for compression |

```
06:        LL = I // // Update image.
07:     end for h
08:     F(t, 1: 11780) = conc(F¹, F², F³, F⁴); //Merge the extracted features.
09: end for t
10: Normalize features.
11: Apply NCA
12: Choose top 1000 features.
13: Feed the chosen 1000 features to SVM and obtain results.
```

More details about the presented model are explained in sub-sections step by step.

### 2.3.1. Preprocessing

Preprocessing is the first phase of the recommended model. By deploying the used multilevel DWT decomposition based preprocessing, pyramidal structure is constructed. By deploying DWT, first, second and third levels (LL1, LL2 and LL3) decomposed images are obtained.

*Step 1:* Employ 3-level DWT to colored skin cancer image using daubechies 4 (db4) mother wavelet function.

$$[LL^1, LH^1, HL^1, HH^1] = DWT(I) \tag{1}$$

$$[LL^k, LH^k, HL^k, HH^k] = DWT(LL^{k-1}), k \in \{2,3\} \tag{2}$$

Herein, $LL^k, LH^k, HL^k, HH^k$ are k[th] level low-low, low-high, high-low and high-high wavelet coefficients respectively.

### 2.3.2. Feature generation using ensemble DarkNet and hand-crafted generators

The features are generated using the raw skin cancer images and LL sub-bands of them. Two types features are generated in here and they are named deep and textural features. Deep features are generated using two pre-trained DarkNets and they are named DarkNet19 and DarkNet53. These networks are generated 2000 features from each image. Textural features are extracted using LBP and LPQ extractors. The used skin cancer images have three layers (R, G and B), hence, these histogram-based textural feature generators are deployed to each channels. 256 and 59 features are extracted using LPQ and LBP consecutively. Thus, 315 features are generated from each layer and 315 x 3 =945 features are generated from each images. In total, 2000 deep features and 945 hand-crafted features are generated and 2945 are obtained from each level. This pyramidal model uses four level, hence, 2945 x 4 =11,780 features are created.

*Step 2:* Generate deep features using pre-trained ensemble DarkNet.

$$df^1 = conc(DarkNet19(I), DarkNet53(I)) \tag{3}$$

$$df^{k+1} = conc\big(DarkNet19(LL^k), DarkNet53(LL^k)\big), k \in \{1,2,3\} \qquad (4)$$

Herein, $df^k$ represents $k^{th}$ deep features with a length of 2000.

**Step 3:** Extract textural features using the LBP and LPQ feature generators. These descriptors are defined applied to R, G and B layers of the image.

$$tf^1 = conc\big(LPQ(I), LBP(I)\big) \qquad (5)$$

$$tf^{k+1} = conc\big(LPQ(LL^k), LBP(LL^k)\big), k \in \{1,2,3\} \qquad (6)$$

where $tf^k$ represents $k^{th}$ deep features with a length of 945.

**Step 4:** Merge the generated deep and textural features and obtain merged features ($mf$) with a length of 11,780.

$$mf = conc(df^j, tf^j), j \in \{1,2,3,4\} \qquad (7)$$

### 2.3.3. Feature selection

A basic threshold based feature elimination and NCA selector based discriminative feature chosen process is used in this research. In the pyramidal feature generation phase, 11,780 features are generated. To improve classification ability of the presented model, feature selection should be employed. Here, two layered selection process is used and they are redundant feature elimination and NCA selector is applied to the selected redundant features. Top (the most informative) 1000 features are selected by NCA.

**Step 5:** Deploy min-max normalization to $mf$.

$$f^N(:,i) = \frac{mf(:,i) - min\big(mf(:,i)\big)}{max\big(mf(:,i)\big) - min\big(mf(:,i)\big)}, i \in \{1,2,\ldots,11{,}780\} \qquad (8)$$

where $f^N$ defines normalized features.

**Step 6:** Calculate summation of each features.

$$tpl(i) = \sum_{j=1}^{3297} f^N(j,i) \qquad (9)$$

where $tpl$ is summation of each feature.

**Step 7:** Eliminate features which summation of feature ($tpl$) is equal to 0.

**Step 8:** Apply NCA to the selected features in Step 7.

**Step 9:** Select top 1000 features using NCA.

### 2.3.4. Classification

SVM classifier is employed to the selected feature matrix with a size of 3297 x 1000. In order to validate results, both hold-out validation (separation ratios are 90:10, 80:20, 70:30, 60:40,

50:50) and ten-fold cross-validation are used. The used SVM classifier is a polynomial order kernel scaled classifier and is called Cubic SVM. This problem is a binary classification problem, hence, only box-constraint level (C) is selected as one.

***Step 10:*** Classify the chosen top 1000 features using Cubic SVM classifier.

## 3. Experimental results

The used skin cancer image dataset contains 3297 colored images and these images are stored as JPG. To develop the proposed pyramidal fused feature generation and NCA selector based model, MATLAB 2020b programming environment is used. The attributes of the used computer are given as follows. The used computer has 48 GB RAM, Intel i9 9900K 3.60 GHz CPU and 64 bit Windows 10.1 pro operating system. The pre-trained DarkNet19 and DarkNet53 was set up to MATLAB and the calculated weights using ImageNet were used for implementing transfer learning. By using DarkNet19 and DarkNet53 in transfer learning mode, features were generated from avg1 and conv53 layers of these network respectively. Furthermore, LBP and LPQ feature generation function were applied to each layer of the images of the constructed pyramid using DWT. Top 1000 features are selected in selection phase and SVM classifies these features using six validations techniques. MATLAB (2020b) classification learner toolbox is used for implementing SVM. Sensitivity, specificity, geometric mean, F1-score and accuracies were calculated for each validation method and the results obtained were given in Table 3.

Table 3. The calculated best results (%) of our presented pyramidal ensemble pre-trained DarkNet and hand-crafted features based model using both hold-out and fold-cross validations.

| Metric | Result (90:10) | Result (80:20) | Result (70:30) | Result (60:40) | Result (50:50) | Result (10-fold cross-validation) |
|---|---|---|---|---|---|---|
| Sensitivity | 96.67 | 93.61 | 92.59 | 93.06 | 91.11 | 91.89 |
| Specificity | 94.63 | 92.31 | 92.65 | 89.63 | 90.91 | 91.12 |
| Geometric mean | 95.64 | 92.96 | 92.62 | 91.33 | 91.01 | 91.50 |
| F1-score | 96.13 | 93.61 | 93.20 | 92.29 | 91.72 | 92.22 |
| Accuracy | 95.74 | 93.02 | 92.62 | 91.50 | 91.02 | 91.53 |

Table 3 denoted that our model attained 95.74% classification accuracy using 90:10 hold-out validation. Further, the proposed model reached >90% accuracy for all validation techniques. These results obviously demonstrated robustness of the presented pyramidal ensemble DarkNet and NCA based skin cancer detection model. Moreover, the calculated general (mean ± standard deviation) accuracies were calculated and these were equal to 90.82 ± 1.65, 90.65 ± 1.02, 90.19 ± 0.90, 89.85 ± 0.73, 89.30 ± 0.63 and 91.06 ± 0.22 by applying 90:10, 80:20, 70:30, 60:40, 50:50 and ten-fold cross validation methods respectively.

## 4. Discussions

Accurate and prompt skin cancer detection/classification is one of the most crucial processes for medical centers. Thus, various automated skin cancer classification/detection model have been proposed/developed in the literature. This research introduces new pyramidal fused feature generator (pre-trained ensemble DarkNet and textural features) and NCA based automatic skin cancer classification model. SVM (it is a shallow classifier) is used to classify the selected features and six validation techniques are used to obtain results. By using these six-validation techniques, the confusion matrices calculated are shown in Figure 6.

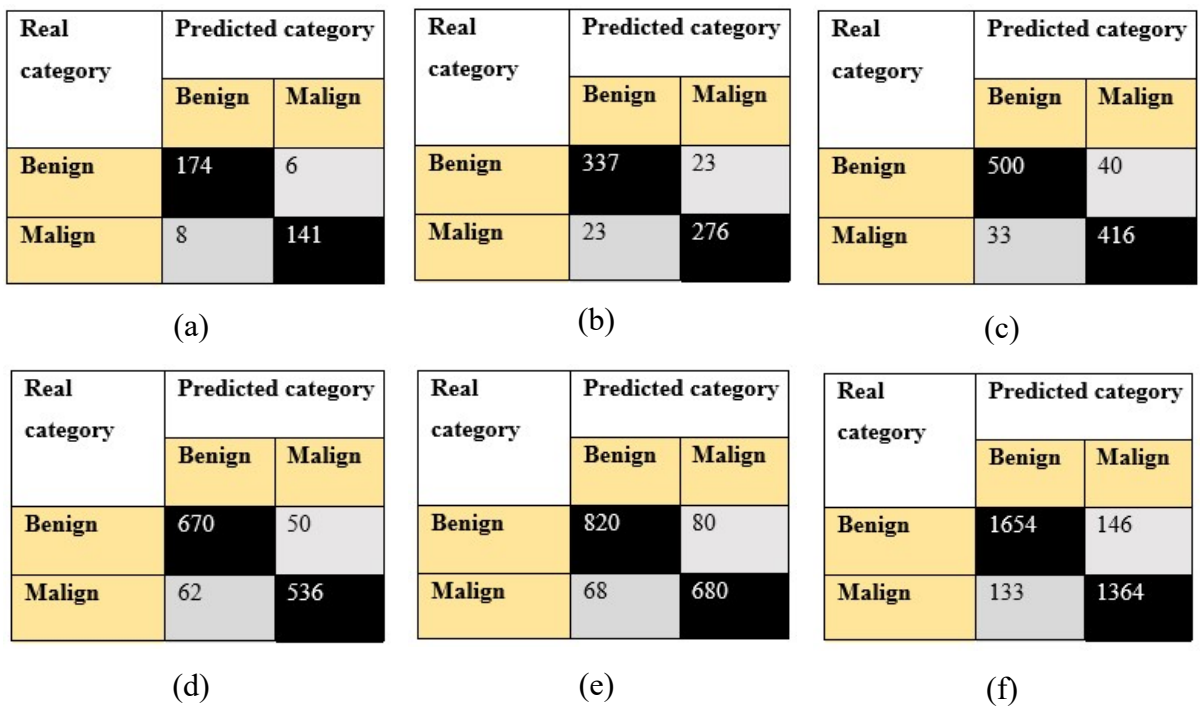

Fig. 6. The confusion matrices calculated with six validations techniques (a) 90:10 (b) 80:20 (c) 70:30 (d) 60:40 (e) 50:50 (f) 10-fold cross-validation.

To indicate success of the introduced model, the presented model is compared to other state-of-art methods and the calculated results are listed in Table 4.

Table 4. Comparative results.

| Authors | Year | Model | Dataset | Classification accuracy (%) |
|---|---|---|---|---|
| Farooq et al. [41] | 2019 | Two deep learning networks (Inception-V3+MobileNet) | Original dataset | 86.0 |
| Togacar et al. [22] | 2021 | MobileNet + Autoencoder + SVM with 10-fold cross validation | Original dataset | 87.11 |
| Togacar et al. [22] | 2021 | MobileNet + Autoencoder + SVM with 10-fold cross-validation | Autoencoder structured dataset | 87.32 |
| Togacar et al. [22] | 2021 | MobileNet + Autoencoder + SVM with 10-fold cross-validation | Original dataset + Autoencoder structured dataset | 93.54 |
| Togacar et al. [22] | 2021 | MobileNet + Autoencoder + Spking Neural Network with 10-fold cross-validation | Original dataset + Autoencoder structured dataset | 95.27 |
| Our method | | Ensemble DarkNet + textural features + SVM with 10-fold cross validation | Original dataset | 91.54 |
| Our method | | Ensemble DarkNet + textural features + SVM with 90:10 split ratio | Original dataset | 95.74 |

When studies conducted on the same dataset in the literature are examined, it is seen that the classification accuracies are 86% by Farooq [41] and 87.11% by Togacar [22], respectively. Also, in the same study by Togacar [22], Autoencoder was applied on the original dataset. Then the original dataset and Autoencoder structured dataset were combined. Classification was made on this new dataset using MobileNet and Spiking Neural Network and a maximum accuracy of 95.27% was found. In this paper, 95.74% accuracy was achieved by using only the original dataset. Ensemble DarkNet and textural features were used for this process, and a classical method SVM classifier was used. As seen from Table 4, the advantages of the presented pyramidal ensemble DarkNet and textural feature generation based skin cancer classification model:

- A hybrid feature generator is presented in this research and benefits of the both deep features and hand-crafted features are used together.
- The presented model used pre-trained network and simple feature extractors are used together in this model. Therefore, implementation of this model is very simple and it can be implemented for real-world application.
- We reached high classification accuracy (See Table 4) on the original dataset.
- The recommended pyramidal model achieved 95.74% accuracy.

Our future plan is to develop a prompt responded intelligent skin cancer detection model. In this model, images of the skin cancers are acquired using a basic camera. These images are loaded the developed intelligent skin cancer detection model. The schematic denotation of our future project for real world application is shown in Fig. 7.

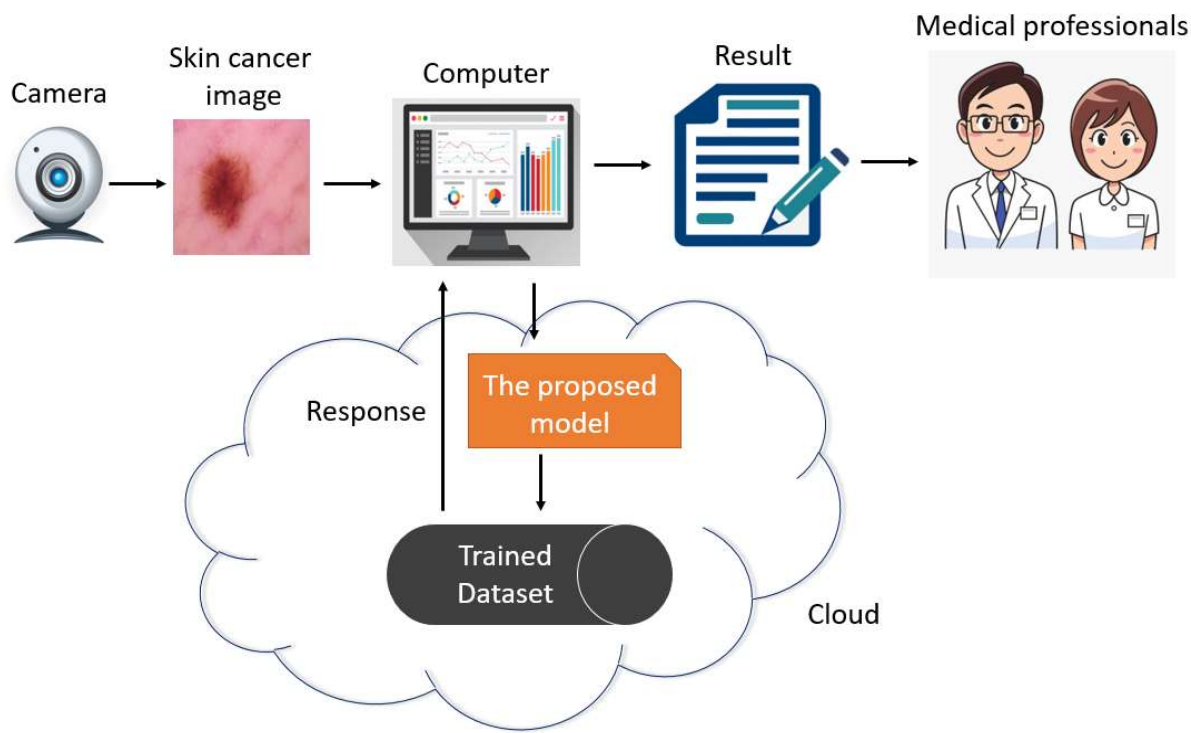

Fig. 7. Graphical explanation of our intended future project.

## 5. Conclusions

Accurate and rapid skin cancer classification/detection is very important and image processing is one of the commonly used way of the detection skim cancer. Therefore, various methods and datasets have been presented in the literature to develop an automatic skin cancer classification model. This model introduces an ensemble DarkNet to extract deep features. Moreover, textural features are used to generate hand-crafted efficient features. A pyramidal feature generation structure is created in this model and skin cancer images are classified using this feature generation structure, NCA and SVM techniques together. This model attained 95.74% accuracy and outperformed. Findings obviously denoted that the presented pyramidal feature generation and NCA based model is ready to use for clinic applications to help physicians for skin cancer diagnosis.